\documentclass{article}

\usepackage[nonatbib,preprint]{nips_2018}

\usepackage[utf8]{inputenc} % allow utf-8 input
\usepackage[T1]{fontenc}    % use 8-bit T1 fonts
\usepackage{hyperref}       % hyperlinks
\usepackage{url}            % simple URL typesetting
\usepackage{booktabs}       % professional-quality tables
\usepackage{amsfonts}       % blackboard math symbols
\usepackage{nicefrac}       % compact symbols for 1/2, etc.
\usepackage{microtype}      % microtypography
\usepackage{graphicx}
\usepackage{epstopdf}
\usepackage{amsmath}
\usepackage{array,multirow}
\usepackage[square,numbers]{natbib}

\title{Proxy Fairness}

\author{
    Maya Gupta  \\
  Google AI \\
  \texttt{mayagupta@google.com} \\
    \And 
  Andrew Cotter  \\
  Google AI \\
  \texttt{acotter@google.com} \\
  \And
    Mahdi Milani Fard \\
  Google AI \\
  \texttt{mmilanifard@google.com} \\
    \And 
 Serena Wang  \\
  Google AI \\
  \texttt{serenawang@google.com} \\
}

\begin{document}

\maketitle

\begin{abstract}
We consider the problem of improving fairness when one lacks access to a dataset labeled with protected groups, making it difficult to take advantage of strategies that can improve fairness but require protected group labels, either at training or runtime. To address this, we investigate improving fairness metrics for \emph{proxy} groups, and test whether doing so results in improved fairness for the true sensitive groups. Results on benchmark and real-world datasets demonstrate that such a proxy fairness strategy can work well in practice. However, we caution that the effectiveness likely depends on the choice of fairness metric, as well as how aligned the proxy groups are with the true protected groups in terms of the constrained model parameters.
\end{abstract}

\section{Introduction}
For moral, legal, and business reasons, it may be of importance to train classifiers that are deemed more \emph{fair}. Different people may have different notions of the ``right'' fairness metric, depending on the context of the problem, and even given the same context, may disagree~\citep{Haidt:2013}. Fairness metrics are typically defined with respect to certain \emph{protected groups} such as gender or ethnicity, but depending on the context, fairness goals may be applicable to many other types of groups, such as small businesses, farmers, bicyclists, country music fans, and so on.

One of the major practical challenges in training machine learning models for group-based policy goals is that we often do not know which examples belong to which groups.  the protected group labels may not be available for privacy or expense reasons, or simply because one did not forsee the need for protected group labels when collecting the training data.

We show that there is some hope for improving fairness metrics even without labeled examples from protected groups, via an experimental investigation of what we term \emph{proxy fairness}: improving fairness metrics on proxy groups as substitutes for the true protected groups.  We show that when the proxy groups are semantically related to the unknown protected groups, in practice fairness improvements on the proxy groups generalize fairly well to the protected groups. More surprisingly, we provide some experimental evidence that being fairer to proxy groups that are seemingly unrelated to the protected groups can also improve fairness metrics for the protected groups. Overall, our experiments suggest that proxy groups can be effective in practice. 

\section{Group-based Fairness and Policy Goals}
Table \ref{tab:goals} details some  different mathematical specifications of group-based fairness goals for classifiers. It is easy to create scenarios where these different fairness goals conflict with each other, both for a given group, and across multiple groups, as well as conflict with the usual goal of maximizing classifier accuracy. See Appendix A in the supplemental for more details on how classifiers can end up doing poorly at these fairness goals, even when no bad intent is at play. In general, we wish to empower machine learning practitioners to achieve policy goals that are measured on groups, and the use of the word \emph{fair} should be considered short-hand for such group-specific goals, and not taken as an endorsement of any particular specific policy goal or stance on fairness.

One category of fairness goals is those that try to better reflect the training data distributions.  For example, the goal of \emph{accurate coverage}  would require that $20\%$ of east-side children and $40\%$ of west-side children were predicted as being eligible for free lunch, if those were the positive label probabilities for each class's training data. Such policy goals  are most relevant when the training data is correctly sampled and labeled for the different groups. 

A second category of goals is where the classifier is asked to disregard some aspect of the training distribution. This usually arises when the training data is not entirely trusted or could be mis-aligned with policy goals, a situation also referred to as \emph{negative legacy} \citep{kamishima:2012}.   For example, a bank might be legally required to give loans at equal rates to both genders.  Aiming for equal average outcomes of the classifier for different groups is also known as \emph{demographic parity} \citep{hardt:2016}, \emph{statistical parity} \citep{fish:2016}, and \emph{equal coverage} \citep{Goh:2016}. 

Two of the goals in Table \ref{tab:goals} reflect an issue we find very important in practice:  how  a \emph{change} to a classifier affects groups.  For example, suppose one invents a new driving test that is more accurate than the current driving test at diagnosing whether senior citizens are safe drivers, then the \emph{not-worse-off} fairness goal would require that the new driving test would not reduce accuracy for teen-aged drivers. Similarly,  \emph{no-lost-benefits} requires a new classifier to classify examples positively from each group at least as often as before (assuming a positive classifiation resulted in a benefit).

%We refer to this practical issue as the \emph{prescient fairness} problem. 

\begin{table*}[t!]
  \caption{Some Group-based Fairness Goals. Notation: $(X,Y,G_1,G_2,\dots,G_K)$ is a random example, where $X$ is the observed feature vector, $Y$ is the observed label, $G_k$ is the indicator for membership of the example in protected group $k$, $\hat{Y}$ denotes the corresponding predicted label. $\hat{Y}_\text{current}$ is a current classifier that $\hat{Y}_\text{new}$ would replace. Variants of these goals can be defined for other metrics such as recall, precision, AUC, etc.} \label{tab:goals}
  \centering
  \begin{tabular}{ll}
    \toprule
 Fairness Goals & Definition \\
        \cmidrule{1-2}
Statistical Parity \citep{Zafar:2017,fish:2016,hardt:2016,Goh:2016} & $ \Pr\{\hat{Y} = 1 | G_k = 1\} = \Pr\{\hat{Y} = 1\} \: \forall k$ \\
Equal Opportunity \citep{hardt:2016}  & $\Pr\{\hat{Y} = 1 | Y = 1, G_k = 1\} = \Pr\{\hat{Y} = 1 | Y = 1\} \: \forall k$\\ 
Equal Odds \citep{hardt:2016} & $\Pr\{\hat{Y} = 1 | Y = y, G_k = 1\} = \Pr\{\hat{Y} = 1 | Y = y\} \:  \forall k$  \\ 
Equal Accuracy  &  $\Pr\{\hat{Y} = Y  | G_k = 1\} = \Pr\{\hat{Y} = Y\} \: \forall k$ \\ 
Accurate Coverage & $\Pr\{\hat{Y} = 1 | G_k = 1\} = \Pr\{Y = 1 | G_k = 1\} \:  \forall k$ \\
Not Worse Off / No Lost Accuracy & $\Pr\{\hat{Y}_\text{new} = Y  | G_k = 1\} \geq \Pr\{\hat{Y}_\text{current} = Y | G_k = 1\} \: \forall k$ \\
No Lost Benefits / No Lost Coverage & $\Pr\{\hat{Y}_\text{new} = 1  | G_k = 1\} \geq \Pr\{\hat{Y}_\text{current} = 1 | G_k = 1\} \: \forall k$ \\
 \bottomrule
 \end{tabular}
\end{table*}

\section{Proxy Fairness}\label{sec:solution}
We propose and investigate creating a set of \emph{proxy groups}  and optimizing for fairness goals for the proxy groups, in the hope that the improved fairness values generalize to the unobservable groups of interest.  We call this \emph{proxy fairness}.

One key question is how to create useful proxy groups. Intuitively, we expect better results if one picks proxy groups that overlap with the true protected groups. For example, for a college admissions classifier, it would be reasonable to use categorical features like applicants' last names as proxy groups for ethnic backgrounds.  

A second key question is how well optimizing for fairness goals on the proxy groups on a training set \emph{generalizes} to improving  the fairness metrics on the true protected groups on a random test set of interest. 

We investigate both of these key questions experimentally in Section \ref{sec:experiments}, and give some theoretical insight into whether a proxy group is useful in Section \ref{sec:theory}.

%Intuitively, we expect better results if one picks proxy groups that overlap with the true protected groups. For example, for a college admissions classifier, one might use categorical features like applicants' last names as proxy groups for ethnic backgrounds. 
%For a speech recognition classifier, one might cluster speakers based on their choice of slang terms, as a proxy for fairness with respect to age. 

\subsection{Optimizing for Group-based Fairness Goals}
Consider a training dataset of $N$ examples $\mathcal D = \{(x_i, y_i, g_{i,1}, g_{i,2}, \dots, g_{i,K})\}_{i = 1, \ldots, N}$, where $x_i \in \mathbb{R}^D$ is the observed feature vector, $y_i \in \{0,1\}$ is the label, and $g_{i,k}$ is the (unobserved) indicator of membership of example $i$ in group $k$, and a test set with the same format is given. We propose using proxy groups $\{\tilde g_k\}$ as substitutes for the true protected groups in any strategy to improve a group-level fairness metric. We thus get augmented dataset $\tilde {\mathcal D} = \{(x_i, y_i, g_{i,1}, g_{i,2}, \dots, g_{i,K},\tilde g_{i,1}, \tilde g_{i,2}, \dots, \tilde g_{i,\tilde K})\}_{i = 1, \ldots, N}$. 

One seeks to produce a classifier that achieves at least some specified value for some choice of fairness metric $J$. For example, if the fairness goal is that the classifier should admit members from the $k$th group to college at least 80$\%$ as often as the overall accept rate, the corresponding fairness metric can be expressed as $J_k(\theta) \leq 0$ where,
\begin{equation*}
J_k(\theta) = 0.8 \frac{ \sum_{\mathcal D}  \hat y_i }{N} - \frac{ \sum_{i=1}^N  g_{i,k} \hat y_i }{ \sum_{i=1}^N  g_{i,k} } . 
\end{equation*}

Our experiments use the post-training fairness strategy where one adds extra parameters to the classifier after training to satisfy the fairness goals \cite{hardt:2016}.  

We analyze the more flexible fairness strategy of explicitly add the fairness goal as a constraint into the training optimization problem \cite{Goh:2016}. Suppose a classifier is given by thresholding a classifier score $f(x_i; \theta)$, where $f \in \mathcal{F}$ for some function class $\mathcal{F}$ parameterized by $\theta \in \mathbb{R}^p$:
\begin{align}
\min_{\theta} &  \sum_{i=1}^N L(f(x_i; \theta), y_i) + R(\theta) \nonumber \\
&\textrm{ s.t. }  J_k(\theta) \leq 0 \textrm{ for all } k =1, \ldots, K  \label{eqn:oracle}
\end{align}

Proxy groups can be substituted for protected groups in other strategies as well. For example, \citet{Zafar:2017} improve fairness metrics by controlling the covariance between each example's membership in multiple groups and the distance to the decision boundary during training. \citet{CaldersVerwer:2010} propose  measuring discrimination as the difference in conditional probabilities, $\Pr\{Y=1 | G_k = 1 \} - \Pr\{Y=1 | G_k = 0 \}$, and a similar measure could be used over proxy groups to help identify potential bias issues. \citet{kamishima:2012} define a prejudice index based on the mutual information of the classification label and protected group membership and penalize such prejudice as a regularization term. \citet{kamiran:2010} propose incorporating information about protected groups into the splitting criteria for decision tree nodes. Russell et al. build causal models for different groups with fairness constraints \cite{Russell:2017}. Many other group-based strategies have been proposed that could be applied with proxy groups.

\subsection{Other Related Work}
Simultaneously with this work, \citet{Kilbertus:2017} also proposed in a separate context using proxy groups in a causal inference framework when interventions on the true protected groups were impossible in practice. They proposed selecting the proxy groups by inferring causal relationships in the underlying data, and then intervening on the proxy groups with the goal of removing problematic causal relationships. In contrast, we make no assumptions about causality in our selection and usage of proxy groups, and suggest proxy groups might be useful with any group-based strategies for improving fairness goals.
 
\citet{Hajian:2013} previously studied the issue that non-sensitive information can be highly correlated with sensitive groups. Their focus was on how such correlations can cause indirect discrimination, and they proposed data pre-processing strategies to ameliorate such problems. Here, we attempt to make constructive use of such correlations by using non-sensitive labels to form proxy groups, and then optimizing model parameters to improve the fairness metrics. 

%\subsection{Fairness Violation}
%Given a dataset... more here, we define the violation of a fairness constraint, or the unfairness, of a classifier parameterized by $\theta$ with respect to %constraint $g$ as $\min(0, \kappa - J(\theta^c, \mathcal{G}_g))$, and the violation of a set of constraints $\{\mathcal{G}_g\}$ as the $\ell_2$ norm of the %violations:  as $\left(\sum_g \left(\min(0, \kappa - J(\theta^c, \mathcal{G}_g))\right)^2\right)^{1/2}$.

\section{What Makes a Good Proxy Group?} \label{sec:theory}
We can be more precise about how the proxy groups should be similar to the protected groups, and we show that choosing poor proxy groups can unnecessarily hurt classifier accuracy, and can even hurt fairness. Fundamentally, a set of proxy groups will be useful if they result in model parameters that are close to the optimal model parameters for the true protected groups, where ``close" is measured by whether the classifier performs similarly on test examples. 

\begin{figure*}[h]
\centering
\begin{tabular}{>{\centering\arraybackslash}b{0.22\textwidth} >{\centering\arraybackslash}b{0.22\textwidth} >{\centering\arraybackslash}b{0.22\textwidth} >{\centering\arraybackslash}b{0.22\textwidth}}
  \multicolumn{2}{c}{\includegraphics[trim={0 10px 0 115px},clip,width=0.45\textwidth]{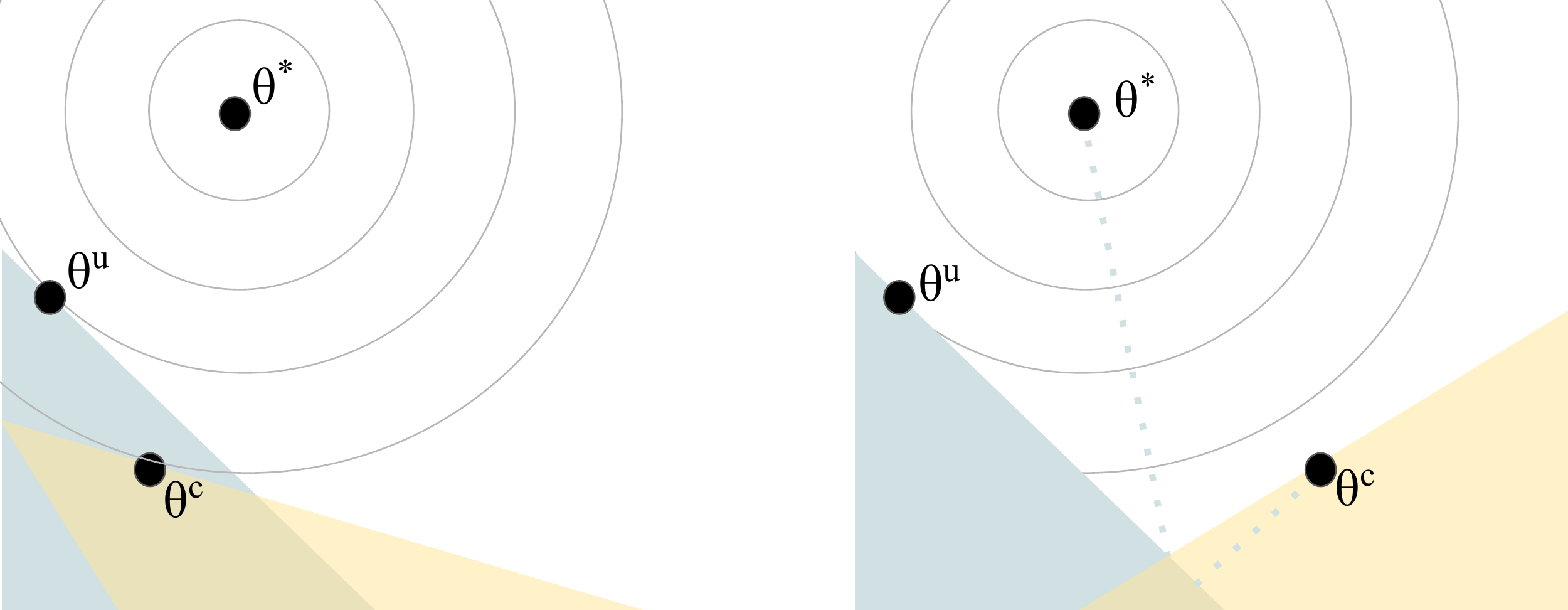}} &
  \multicolumn{2}{c}{\includegraphics[width=0.45\textwidth]{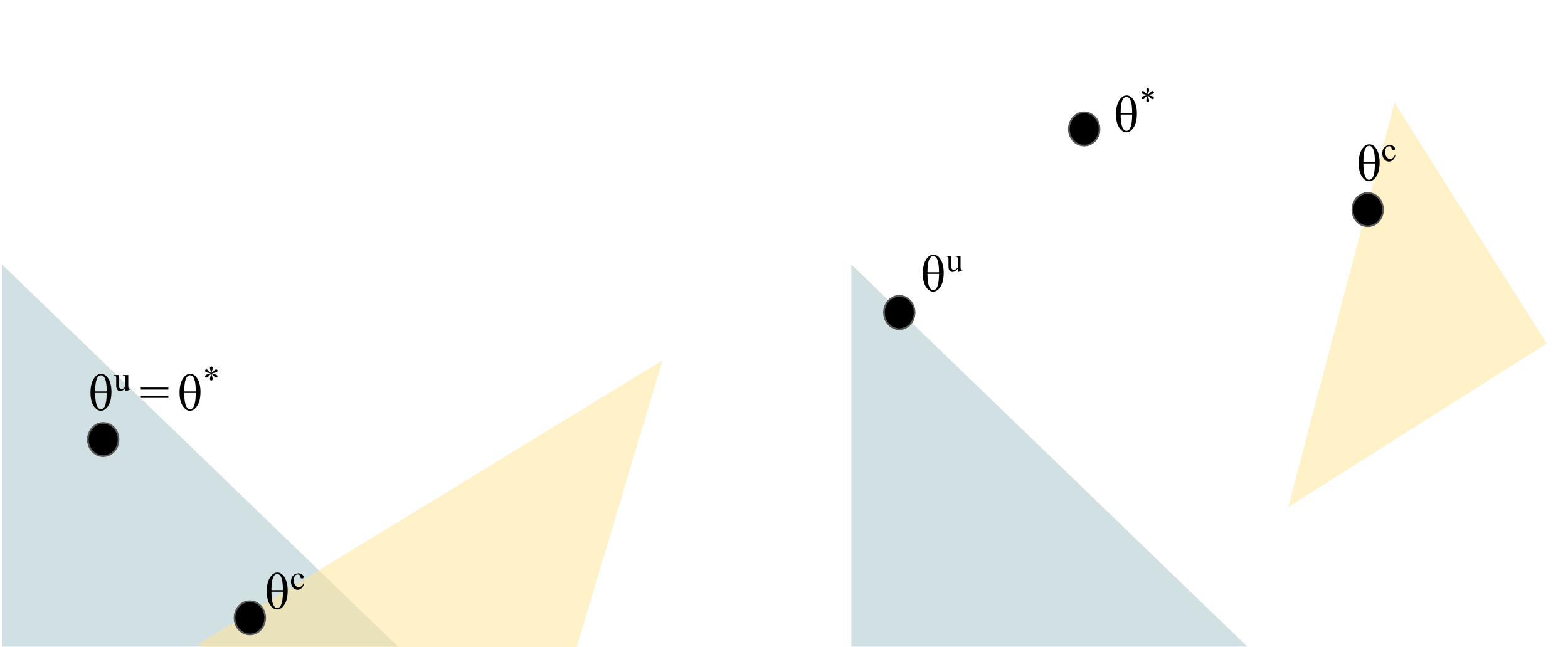}}\\
  (A) & (B) & (C) & (D)
\end{tabular}
\vspace{-1em}
\caption{Illustration of four cases of optimizing classifier parameters with fairness constraints. The blue region defines the fairness constraints for the true groups, and the yellow region shows the fairness constraints for proxy groups. Each $\theta^*$ denotes the classifier parameters that minimize the unconstrained problem, $\theta^u$ minimizes the problem with fairness constraints for the true (but assumed unknown) protected subsets, and $\theta^c$ minimizes the problem with fairness constraints for the proxy subsets, and we assume the fairness constraints form linear inequalities on the classifier parameters (true for standard fairness metrics and linear classifiers). The lemma makes the simplifying assumption that the training objective (\ref{eqn:oracle}) is sufficiently locally symmetric in $\theta$ such that the constrained problems are simply a projection of the unconstrained solution $\theta^*$ onto the constraint set. \textbf{(A)} -  The proxy constraints and the true constraints overlap sufficiently with respect to the unconstrained solution $\theta^*$ that the proxy solution $\theta^c$ is also be a feasible solution to the true fairness constraints, though projecting onto the proxy constraints is less optimal in terms of the objective than it would have been to project onto the true constraints, such that $\theta^c$ incurs a larger loss of accuracy but does satisfy the true fairness constraints.  \textbf{(B)} Here, the feasible regions for the proxy constraints (yellow) and true constraints (blue) do overlap, and thus the proxy-constrained solution $\theta^c$ is closer to the intersection of the true and proxy constraints than $\theta^*$. Then if the fairness metric contours decay with Euclidean distance from the intersection, then the proxy solution $\theta^c$ will have better fairness metrics than the unconstrained solution $\theta^*$.  But the lemma does not promise that the proxy solution $\theta^c$ will be closer to the ideal solution $\theta^u$ than $\theta^*$, and in fact if the intersection of the constraint sets occurs far from $\theta^u$, the proxy solution $\theta^c$ could worsen fairness metrics. \textbf{(C)} Illustrates the special case that the unconstrained solution satisfies the true fairness constraints, and then under the lemma's assumption that the feasible regions for the true constraints and proxy constraints overlap, the proxy constrained solution $\theta^c$ will also satisfy the true constraints, though the projection from $\theta^*$ to $\theta^c$ will incur a cost to the objective (not shown). \textbf{(D)} A worst-case example where the proxy constraint set is a poor approximation as it does not intersect the true fairness constraint set (violating lemma's assumptions). In such cases satisfying the proxy constraints can result in worse fairness metrics as well as worse accuracy than the unconstrained $\theta^*$.
  }
\label{fig:fairnessproxy} 
\end{figure*}

We formalize this in the lemma below for a simplified situation. This lemma suggests one can use a small set of relaxed proxy constraints to safely achieve some gain in protected group fairness, but that a larger set of more restrictive proxy constraints might actually hurt the fairness metrics for the protected groups.  Figure \ref{fig:fairnessproxy} illustrates four example cases.

\textbf{Lemma:} Suppose the true constraint set $\{J_k(\theta) \geq 0\}_{k = 1, \dots, K}$ and the corresponding proxy constraint set $\{\tilde J_k(\theta) \geq 0\}_{k = 1, \dots, \tilde K}$ have a non-zero intersection $\mathcal{I}$ and are both convex sets.  Let $\theta^*$ denote the parameters that minimizes the optimization objective of (\ref{eqn:oracle}) without the constraints $\{J_k(\theta) \geq 0\}_{k = 1, \dots, K}$.  Let $\tilde{\theta}^c$ denote the parameters that minimize the optimization objective subject to the proxy fairness constraints. Suppose the Hessian of the objective is a scaled identity matrix.  Then $\tilde{\theta}^c$ is not further in $\ell_2$ norm to $\mathcal{I}$ than $\theta^*$. \\ 

\textbf{Proof:}
Under the given assumptions, optimizing (\ref{eqn:oracle}) with proxy groups degenerates into a projection of the unconstrained optimizer $\theta^*$ onto the convex proxy constraint set, which cannot increase the distance to the intersection with the convex set of true fairness constraints (see, e.g. Lemma 3.2 of Bauschke and Borwein \cite{bauschkeBorwein:96}).\\

\section{Experiments}\label{sec:experiments}
We investigate the effect of proxy fairness on several real datasets from a large internet services company [name redacted for blind review], and also on the benchmark dataset \emph{Adult}. For each dataset, we impose policy goals only for the purposes of experimentally investigating proxy fairness, and it is not our intent to endorse any specific policy goal.  

For all our experiments, we use the fairness-improving strategy of Hardt et al. \cite{hardt:2016} of taking as given an already-trained classifier $f(x)$, and then adding an additional additive parameter $\beta_k \in \mathcal{R}$ for each proxy group, and optimizing that added parameter to improve the fairness metric on the same training data used to train $f(x)$, resulting in the proxy fairness corrected classifier $\tilde f(x) = f(x) + \sum_k \beta_k \tilde g_k$.

To verify the gains in the fairness metric are not simply from having the additional degrees of freedom $\{\beta_k\}$ for the classifier, we compared to optimizing the $\{\beta_k\}$ parameters for accuracy. The same pre-trained classifier $f(x)$ is used for all experiments for a given dataset.

We restrict our focus to fairness goals of \emph{accurate coverage} \emph{equal opportunity} (see Table 1). We report the corresponding unfairness metrics:

\noindent \textbf{Mean Coverage Error: } $\frac{1}{K} \sum_{k} \left \vert \frac{\sum_i (\hat y_i -y_i) g_{i,k}}{ \sum_i g_{i,k} } \right \vert$  \\

\noindent \textbf{Mean Equal Opp. Error: } $\frac{1}{K} \sum_{k} \left \vert \frac{ \sum_i \hat y_i y_i g_{i,k}}{ \sum_i y_i g_{i,k}} - \frac{\sum_i \hat y_i y_i }{ \sum_i y_i} \right \vert$

All reported fairness metrics are calculated on the true protected groups of interest.   

\subsection{Case Study: Business Size}
The classifier decides whether to recommend a particular local business to a user. We test the \emph{accurate coverage} fairness goal with respect to the size of the business, grouped as small, medium, or large (S/M/L), so that businesses are accurately proportionately recommended regardless of size. For proxy groups, we use the number of user-submitted photos for a business, grouped into 3 uniform quantiles. There were $N = 5,242,880$ training and $1,048,576$ testing examples, and the underlying classifier is a nonlinear ensemble of $20$ x $8$-dimensional lattices \citep{canini:2016} trained on $D=20$ features.  

Table \ref{tab:hotels} gives the fairness results in terms of the error in predicted vs. true coverage averaged over the true S/M/L groups.  The results show that enforcing \emph{accurate coverage} on the proxy groups (\# photos) on the train set does improve the test set fairness metric by 1/3, without damaging test accuracy. Had we post-shifted on the true protected groups (S/M/L), the fairness improvement on the test set would have been much better (almost $100\%)$, though traded-off for test accuracy. The comparisons to post-shifting for accuracy (rather than for the fairness goal) confirm that the gains are not simply from adding the three additional free parameters to the classifier.

\begin{table}[h!]
\vspace{-0.5em}
\caption{Results for business size with \# photos as a proxy.}
\label{tab:hotels}
\begin{center}
\begin{tabular} {llllll}
\toprule
Post & Post & Train &  Test & Mean   & Mean  \\
 Shift & Shift & Acc. & Acc. & Train & Test  \\
Feature& Obj. &  &  & Cov. & Cov.  \\
&  &  &  & Error & Error  \\
\midrule
None  & None     & 56.65 & 56.51 & 9.11 & 9.28  \\ 
\# Photos  & Fair. & 56.58 & 56.43 & 6.41 & 6.63 \\ 
\# Photos  & Acc. & 56.67 & 56.52 & 8.97 & 9.17 \\
 S/M/L & Fair. & 56.49 & 56.30 & 0.0 & 0.31 \\ 
S/M/L & Acc. & 56.67 & 56.50 & 10.53 & 10.72  \\ 
\bottomrule
\end{tabular}
\end{center}
\vspace{-0.5em}
\end{table}

\subsection{Case Study: Provider ID}
The classifier decides whether to show a candidate result to a user, where the candidate result comes from one of $500$ distinct providers. We test the \emph{accurate coverage} fairness metric across the $500$ providers, which are very unevenly represented in the dataset. 
For proxy groups, we place the providers into $K$ size buckets based on the number of examples belonging to that provider in the training dataset.  We vary $K$  to analyze the effect of more proxy groups. There are $N = 249,174$ training and $62,182$ testing examples, and the classifier is a non-linear lattice model \citep{GuptaEtAl:2016} trained on $D=5$ features. 

Table \ref{tab:LQ_full} reports the mean coverage error averaged over the 500+ providers. First, we note that post-shifting on the true protected groups for this fairness metric does reduce the coverage error from $16.11$ to $6.31$, but at a hefty cost to test accuracy. The controlled experiment uses the extra degrees of freedom to optimize the classifier for training accuracy, and does raise test accuracy but also increases the error in accurate-coverage. 
Post-shifting for the fairness metric on the proxy groups does help, with the finer-grained (and more accurate) proxy groups providing lower test mean coverage error on the true protected groups, at the cost of worse test accuracy. 

%However, we caution that using a large number of proxy groups is only be useful if one has enough data to avoid over-fitting to the training data when constraining for fairness. We have included a cautionary experiment in the supplemental materials to demonstrate the possibility of over-fitting fairness constraints further, where even using the true groups (not proxies) has an overfitting problem.

\begin{table}[h!]
\vspace{-0.5em}
\caption{Results for Providers with $K$ size buckets as proxies.}
\label{tab:LQ_full}
\begin{center}
\begin{tabular} {llllll}
\toprule
Post & Post & Train &  Test & Mean   & Mean  \\
 Shift & Shift & Acc. & Acc. & Train & Test  \\
Feature& Obj. &  &  & Cov. & Cov.  \\
&  &  &  & Error & Error  \\
\midrule
None           & None     & 67.29 & 66.78 & 15.33 & 16.11  \\ 
$K=2$    & Fair. & 67.15 & 66.58 & 15.17 & 15.90 \\  
$K=50$  & Fair. & 66.68 & 66.03 & 14.26 & 14.96  \\  
$K=100$ & Fair. & 66.45 & 65.90 & 12.70 & 13.94  \\  
Provider          & Fair. & 65.79 & 65.33 & 0.0 & 6.31 \\
Provider          & Acc. & 69.12 & 67.49 & 18.15 & 18.75  \\  

\bottomrule
\end{tabular}
\end{center}
\vspace{-0.5em}
\end{table}

\begin{table*}[h!]
\caption{Results for Adult dataset for the fairness goal of accurate coverage.}
\label{tab:Adult_accCov_full}
\begin{center}
\begin{tabular} {llllllll}
\toprule
Post-Shift & Post-Shift & Train & Test & Mean Train & Mean Test  & Mean Train & Mean Test \\
Feature & Obj. & Acc. & Acc. & Cov. Err. & Cov. Err.  & Cov. Err. & Cov. Err. \\
 &  &  &  & Gender &  Gender & Race & Race \\
\midrule
None & None            & 85.80 & 85.80 & 3.13 & 2.82 & 2.15 & 3.73 \\
Gender & Fairness      & 85.39 & 85.19 & 0.0 & 0.48 & 0.76 & 2.23 \\ 
Race & Fairness        & 85.42 & 85.21 & 1.33 & 1.47 & 0.0 & 2.14 \\ 
Gender & Accuracy      & 85.95 & 85.73 & 3.28 & 2.95 & 2.99 & 3.77 \\
Race & Accuracy        & 85.97 & 85.79 & 4.03 & 3.62 & 3.57 & 4.62 \\
\bottomrule
\end{tabular}
\end{center}
\vspace{-0.5em}
\end{table*}

\begin{table*}[t]
\begin{small}
\caption{Results for Adult dataset for the fairness goal of equal opportunity. }
\label{tab:Adult_eqnOpt_full}
\begin{center}
\begin{tabular} {llllllll}
\toprule
Post-Shift & Post-Shift & Train & Test & Mean Train & Mean Test  & Mean Train & Mean Test \\
Feature & Obj. & Acc. & Acc. & Eq Opp Error & Eq Opp Error   & Eq Opp Error  & Eq Opp Error  \\
 &  &  &  & Gender &  Gender & Race & Race \\
\midrule
None & None            & 85.80 & 85.80 & 6.69 & 6.04 & 6.06 & 8.66 \\
Gender & Fairness      & 85.39 & 85.19 & 0.09 & 0.43 & 7.66 & 8.53 \\ 
Race        & Fairness & 85.42 & 85.21 & 6.21 & 5.58 & 0.60 & 4.71 \\
Gender & Accuracy      & 85.95 & 85.73 & 2.88 & 2.79 & 7.87 & 8.14 \\
Race & Accuracy        & 85.97 & 85.79 & 6.49 & 5.40 & 9.39 & 10.86 \\ 
\bottomrule
\end{tabular}
\end{center}
\end{small}
\vspace{-0.5em}
\end{table*}

\subsection{Adult Benchmark Dataset: Gender and Race }
Each example in the benchmark UCI Adult dataset  \citep{Lichman:2013} is labeled with one of two genders, and one of five races.  We removed both the gender and race features from the classifier's inputs for all these experiments (however, note that there is still a marital status feature with categories \emph{husband} and \emph{wife} that correlate strongly with gender for married examples). We used the 2 genders as proxy groups for improving the fairness metric with respect to the 5 race groups, and used the 5 race groups as proxy groups  for improving the fairness metric with respect to the 2 gender groups. 

Table \ref{tab:Adult_accCov_full} shows the results on the Adult dataset for the \emph{accurate coverage} fairness metric. We were surprised to see that race/gender proxy groups substantially improved the test fairness values on the other protected group.

However, Table \ref{tab:Adult_eqnOpt_full} shows the results of a second experiment with this dataset, where the fairness metric is changed to \emph{equal opportunity}. For this metric, the race/gender proxy groups barely help.  In general, we find  \emph{equal opportunity} to be a harder and more fragile fairness goal, we suspect because it depends on the protected groups' label distributions.

\subsection{Case Study: Business Entity Resolution}
This classifier decides  whether two business descriptions are about the same business (also called \emph{entity resolution}). We test the \emph{accurate coverage} fairness goal for $16$ different international regions, which are very unevenly represented in the dataset.  We form two proxy groups depending on whether the business is known to be a chain or not. We expect different countries to have different frequencies of known chain businesses, but this is expected to be a poor proxy for the true protected groups. There are $N = 15,415$ training and $3,854$ test examples, all IID, and the classifier is a non-linear lattice model~\citep{GuptaEtAl:2016} trained on $D = 9$ features.

Table \ref{tab:mapfacts_full} shows that despite  using only two  proxy groups to improve fairness values on sixteen countries, and despite the proxy groups being only loosely correlated with country, the proxy fairness does reduce the true groups' test coverage error (improving this fairness goal) by $10\%$, but also causes a notable hit to accuracy.

\begin{table}[h]
\caption{Results for business entity resolution with 16 regions as the true protected groups, and chain business proxy groups.}
\label{tab:mapfacts_full}
\begin{center}
\begin{tabular} {llllrr}
\toprule
Post & Post & Train &  Test & Mean   & Mean  \\
 Shift & Shift & Acc. & Acc. & Train & Test  \\
Feature& Obj. &  &  & Cov. & Cov.  \\
&  &  &  & Error & Error  \\
\midrule
None    & None     & 84.75 & 84.82 & 3.71 & 4.49 \\
Country & Fairness & 85.05 & 84.85 & 0.0 & 2.57 \\  
Chain   & Fairness & 84.68 & 84.43 & 3.27 & 4.14 \\ 
Country & Accuracy & 85.62 & 85.24 & 3.65 & 3.25 \\  
Chain   & Accuracy & 85.12 & 84.98 & 7.09 & 7.92 \\ 
\bottomrule
\end{tabular}
\end{center}
\vspace{-0.5em}
\end{table}

 \subsection{Case Study: Region and Language}
 The classifier decides whether a candidate result should be added to a list of recommendations.  We test the \emph{accurate coverage} fairness goal. We consider two different possible sets of protected groups:  24 different languages, or 35 international regions, both of which are very unevenly represented in the dataset, with two-thirds  of examples being from the United States, and three-quarters of examples being in English. We experiment with either language or region as the true groups, and  the other categorization as the proxy groups.  Language is mostly a super-set of regions, but some regions do have multiple languages, so this is not a strict superset/subset relationship (for example, examples from India are either in English or Hindi). There are $N = 62,220$ training and $7,919$ testing examples, and the base classifier is a non-linear lattice model \citep{GuptaEtAl:2016} trained on $D=10$ features.

 Table \ref{tab:LuluThresholding} shows the results for two sets of experiments. For the first set we use post-shifting for all $35$ countries or $24$ languages (i.e. one shifting parameter for each country/language). Since some of the countries and languages have a very small set of training/testing examples, post-shifting for all 35 countries or 24 languages causes over-fitting: the train fairness errors decrease but the test fairness error gets worse! This over-fitting happens with both the true groups and the proxy groups for this problem.  

 In the second set of experiments, we avoid the over-fitting issue by not post-shifting for countries for which we do not have enough data to generalize well to the testing distribution: we only apply the post-shifting to the top $10$ countries/languages for each of which we have at least $300$ training examples. This helps the over-fitting and the proxy groups achieve roughly half the fairness metric improvement of the true groups, at roughly the same accuracy.  The gains even with the true protected groups is small though, because most of the coverage error occurs for the smaller languages/countries that do not get the post-shifting correction.

 \begin{table*}[h]
 \caption{Results for Region/Language Groups, where either 35 regions, or 24 languages are taken to be the protected groups, with the other category acting as the proxy groups. The second set uses 10 biggest groups for each.}
 \label{tab:LuluThresholding}
 \begin{center}
 \begin{tabular} {llrrrrrr}
 \toprule
 Post-Shift & Post-Shift & Train & Test & Mean Train & Mean Test & Mean Train & Mean Test \\
 Feature & Obj. & Acc. & Acc. & Cov. Err. & Cov. Err. & Cov. Err. & Cov. Err. \\
 & & & & Language &  Language & Country & Country \\
 \midrule
 None     & None     & 64.35 & 64.92 & 12.05 & 24.02 & 15.31 & 21.85 \\ 
 \midrule
 Lang.-24 & Fairness & 64.53 & 64.85 & 0.0 & 25.31 & 6.25 & 23.86 \\ 
 Country-35  & Fairness & 64.55 & 64.83 & 0.28 & 25.25 & 0.0 & 24.14 \\  
 Lang.-24 & Accuracy & 64.80 & 64.28 & 9.89 & 31.05 & 11.31 & 27.17  \\
 Country-35  & Accuracy & 64.84 & 64.38 & 8.81 & 30.80 & 11.02 & 27.43 \\ 
 \midrule
 Lang.-10 & Fairness & 64.51 & 64.92 & 9.77 & 22.84 & 12.80 & 20.81 \\ 
 Country-10  & Fairness & 64.47 & 64.92 & 10.72 & 23.83 & 14.17 & 21.23 \\ 
 Lang.-10 & Accuracy & 64.73 & 64.29 & 11.19 & 23.94 & 14.05 & 21.39 \\
 Country-10  & Accuracy & 64.73 & 64.44 & 12.03 & 24.33 & 15.54 & 22.46 \\  
 \bottomrule
 \end{tabular}
 \end{center}
 \end{table*}

\section{Conclusions}
We have raised the practical problem of lack of protected group labels, and  investigated the solution of proxy fairness.  Our experimental results were consistent with the intuition that creating proxy groups that are semantically related to the true protected groups will improve fairness metrics.  However, we were surprised to find that even with weakly related proxy groups, such as gender/race and chain/region, we saw improvements to test fairness metrics on the true groups for certain fairness goals. Lastly, we  demonstrated the potential for overfitting to fairness metrics and side effects on classifier accuracy.  

Our experiments were limited to the fairness goals of \emph{accurate coverage} and \emph{equal opportunity}.  We hypothesize that fairness goals correlated with classifier accuracy are also more likely to be amenable to proxy fairness. For example, \emph{equal opportunity} generally correlates positively with classifier accuracy.  See Appendix B for a deeper discussion of how different fairness goals interact with optimizing classifier accuracy.  We note that many (but not all) fairness metrics can also be improved by simply improving the overall accuracy of the classifier, which should be tested as a control for any  proposal to increase fairness metrics (as we did here).  

A set of proxy groups is useful if optimizing for fairness for the proxy groups affects the model parameters in the same way as optimizing for fairness on the true groups. That leads us to hypothesize that proxy fairness will generally be more effective with less flexible models, but the actual effect of model complexity on the effectiveness of proxy fairness remains an open question. 

%Our theoretical analysis shows that a bad choice of proxy groups and very aggressive fairness constraints can hurt fairness metrics and accuracy. 

\clearpage
\newpage

\section{Appendix A: Challenges in Training Fairer Classifiers}\label{sec:causes}
If the given training data was labeled or sampled in a way that is at-odds or mis-aligned with one's specific fairness goals, then of course those fairness metrics may suffer; the ``Catalog of Evils''  in \citet{dwork:2012} specifies a number of such bad behaviors. However, even in the absence of bad actors or bad intentions,  we have found there are challenges to training fairer classifiers. Below we list the major non-adversarial issues that we have identified so far that can make achieving the goals in Table 1 challenging, falling into two categories: (i) ways that training fails to achieve fairness goals even if there are no problems inherent in the training data, (ii) ways that training data can be sampled or labeled suboptimally despite lack of bad intentions. 

Awareness of these issues can help identify and pro-actively address fairness goals for machine-learned systems, both in the prescient fairness setting or with access to labeled protected class examples.  Considering specific causes also highlights the subtlety and complexity of fairness issues, and brings to light that different fairness goals (e.g. as defined in Table \ref{tab:goals})  
may conflict.  Many of these issues are also challenges for human judgment and decision-making, and may in fact be easier to identify and combat in machine-learned systems than in human decision-making processes. 

\subsection{Challenges in Training Fairer Classifiers}

\noindent \textbf{Tyranny of the Majority:} Empirical risk minimization optimizes for accuracy averaged over the training examples, which is generally not the same thing as accuracy averaged over groups or other group-wise metrics. This risk is greater with less flexible models, or when there are no features that let the model differentiate between groups.

\noindent \textbf{Data Disparity:} If there are relatively few samples for a group, it is statistically more challenging to be as accurate for that group. For example, a restaurant recommendation system may have a harder time accurately predicting whether someone from Montana will like an Indonesian restaurant compared to a Chinese restaurant, because there are fewer Indonesian restaurants in Montana.

\noindent \textbf{Lack of Features:} The classifier features may not be sufficient to achieve a desired fairness goal. For example, different acoustic information may be needed to reach the same accuracy at speech recognition for tonal languages as non-tonal languages. Features that could be useful in achieving (or measuring) fairness goals are often not collected or used in order to protect privacy, producing a privacy-fairness trade-off. Analogous to Bayes error, for a given fairness goal and a given classification error rate and feature set we refer to the maximally achievable value  as the  \emph{Bayes fairness} for that goal.

\noindent \textbf{Lack of Function Class Flexibility:} The function class may not be flexible enough to achieve the desired fairness. 

\noindent \textbf{Differential Hardness:} Examples from one group may be statistically harder to classify than others. For example, it is more difficult to judge the weight of people who wear loose clothes compared to people who wear tight clothes. 

\noindent \textbf{Parameter Regularization:}  Parameters corresponding to some groups may get regularized harder. In particular, with popular additive regularizers like lasso or ridge, the empirical cost to regularizing parameters supporting rarer groups may be small compared to the empirical cost of regularizing parameters corresponding to more common groups. For example, consider a logistic regression classifier for movie ratings where the presence of each actor in a movie is a binary feature with a corresponding parameter, and suppose a lasso regularizer is used to encourage sparsity. Actors with fewer movies may end up more regularized, implementing the \emph{Matthew effect}, in which credit for group work is overly awarded to the more prominent member of the group \citep{Merton:1968}.

\noindent \textbf{Uneven Misclassification Costs:} A classifier may prioritize accuracy on groups with higher misclassification costs.  

\noindent \textbf{Simpson's Paradox:} A classifier may achieve lower aggregate fairness values than when sliced appropriately. The classic example is when women were admitted to Berkeley's graduate programs at overall lower rates than men in 1973, because they were applying to more selective departments on average \citep{Bickel:75}. 

\subsection{Challenges in Obtaining Fairer Training Data}

\noindent \textbf{UI Bias (First-Come, First-Served):}  UI (user-interface) bias occurs when examples are presented in some order for labeling. For example, if one trains on which movies a user watches from a list of recommendations, past highly-placed recommendations will tend to get more watches. The first-come first-served issue was perhaps first identified by Merton \cite{Merton:1968}, who referred to it as the \emph{41st chair} problem, based on the French Academy only have 40 members, and as a result many preeminent thinkers, such as Descartes, were unable to become positively-labeled examples for the French Academy, and French Academy membership becomes inadvertently biased against newer fields of knowledge. The modern version of this problem is in web applications that use past popularity as a signal, and then fresh or new examples are at a disadvantage to even get a chance to be shown to users/raters to be labeled.

UI Bias can cause small biases in the original ordering to spiral into severe biases due to positive feedback. 

\noindent \textbf{Under Sampled:} If the training set has relatively few examples from a protected group, then the classifier will be more subject to problems of \emph{tyranny of the majority} and \emph{data disparity} detailed above. 

\noindent \textbf{Prominence Labeling Bias:} Less prominent/rarer classes are more likely to be incorrectly labeled, due to lack of knowledge or familiarity on the part of the labeler; also known as the \emph{availability heuristic} \citep{kahneman:2011}. For example, most people will label an image of a porpoise as a dolphin. 

\noindent \textbf{Stale Data:} If the sampling and label distribution over time changes and training data is used from a wide time range, training data may be biased, even if it was not biased when collected. For example, the proportion of female doctors in the US has changed dramatically over the last 50 years. 

\noindent \textbf{Social Pressure:} Biased labeling can occur because people either feel social pressure to label a certain way or because they are imagining answers for others, and such biases may differ between groups. For example, this can happen when labeling whether you find an image offensive, or reporting one's weight or age.  

\noindent \textbf{Irrelevant Priors (Halo effect):}  The \emph{halo effect} is the use of irrelevant priors in labeling \citep{thorndike:1920}\citep{kahneman:2011}; for example, labeling a man as kind because he is handsome. 

\noindent \textbf{Other Cognitive Biases:} Other cognitive biases \citep{kahneman:2011} can play a role in bad sampling or bad labeling.

\section{Appendix B: Interactions Between Fairness Goals and Empirical Risk} \label{sec:conflict}
In this section we summarize our findings from many experiments (not reported in the paper) trying to optimize different fairness metrics discussed in Table \ref{tab:goals} in real-world classification settings, using the strategies of post-shifting \citep{hardt:2016} and constrained optimization \citep{Goh:2016}.  We have found that certain fairness metrics are harder to improve than others, depending on how well the goal aligns with empirical risk minimization.  Of course, if one does not trust the training labels, then classifier correctness may not be an important goal. But we found in practice that in the worst cases, fairness goals are only achieved with degenerate classifiers, such as a random classifier. In the best cases, adding fairness constraints can act as a form of regularization that slightly improves generalization and testing accuracy. 
 
\textbf{Equal Accuracy:}  Optimization for this goal tends to be dominated by the group that is hardest to classify, which pulls down the accuracy of other groups towards the worst accuracy, leading to poor or even degenerate classifiers. The risk of poor solutions tends to worsen the more protected groups are included. A thought experiment helps see why this goal is so difficult: suppose one had a classifier that perfectly achieved the goal of equal accuracy across groups. Can one increase the classifier's accuracy while still maintaining equal accuracy? Only if the accuracy increase is uniform across the classes, which is difficult for the empirical risk minimization to do, particularly if we use a stochastic gradient method. 

\textbf{Demographic Parity:} This goal rewards the classifier for classifying each group as positive at the same rate (that is, equal coverage), and can generally be improved at a reasonable trade-off for classifier correctness (see e.g. plots in \citet{Goh:2016}). Re-visiting our thought experiment: suppose one had a classifier that satisfied this goal, for example such that the positive classification rate was $80\%$ for every group. If the groups are somewhat separable in the feature space, then it may be possible to increase the accuracy of this classifier, by changing within a group which of the examples are labelled positively. However, if the true positive rates for the groups are different, the ability to increase the classifier accuracy will be limited. 

\textbf{Not Worse Off:} This goal requires a new classifier to not make any group less accurate than with the current classifier.  Suppose one had a new classifier that satisfied this goal. Then one could increase its accuracy for any random feature vector $X$ and still satisfy the goal. Thus this is generally the easiest fairness goal to optimize without hurting classifier correctness. 

\begin{figure*}[t!]
\centering
\vspace{-0.5em}
\includegraphics[width=0.32\textwidth]{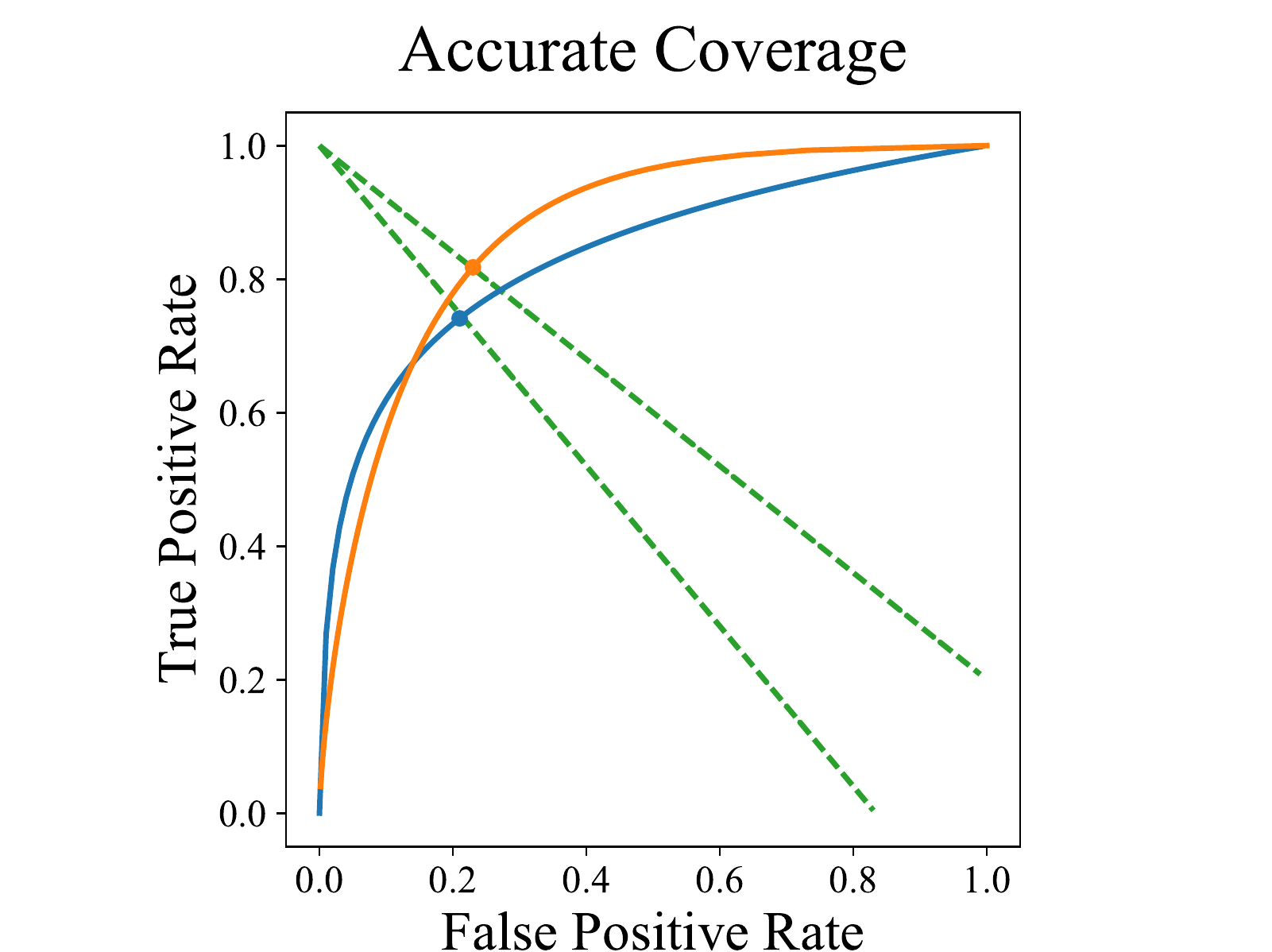}
\includegraphics[width=0.32\textwidth]{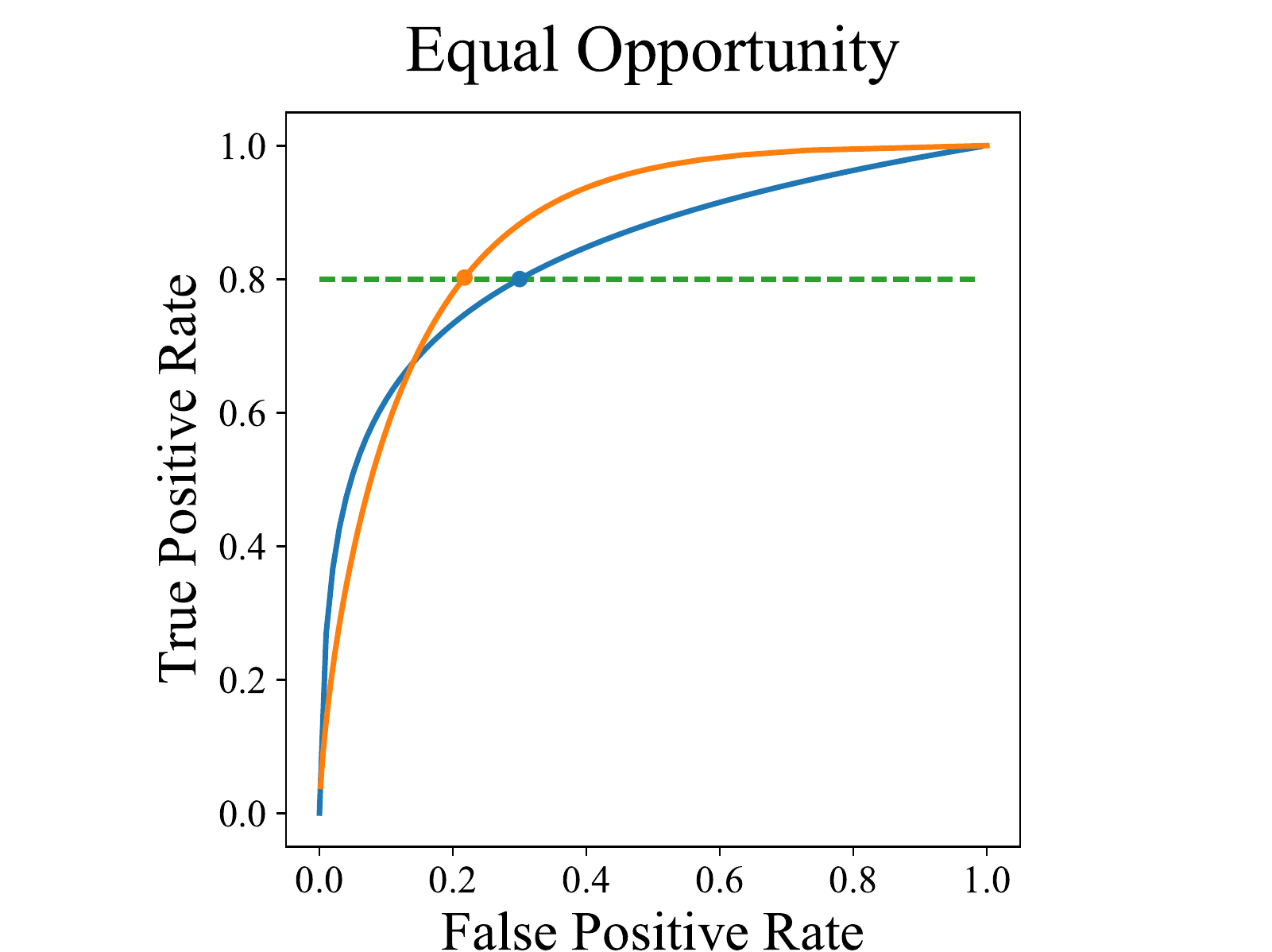}
\includegraphics[width=0.32\textwidth]{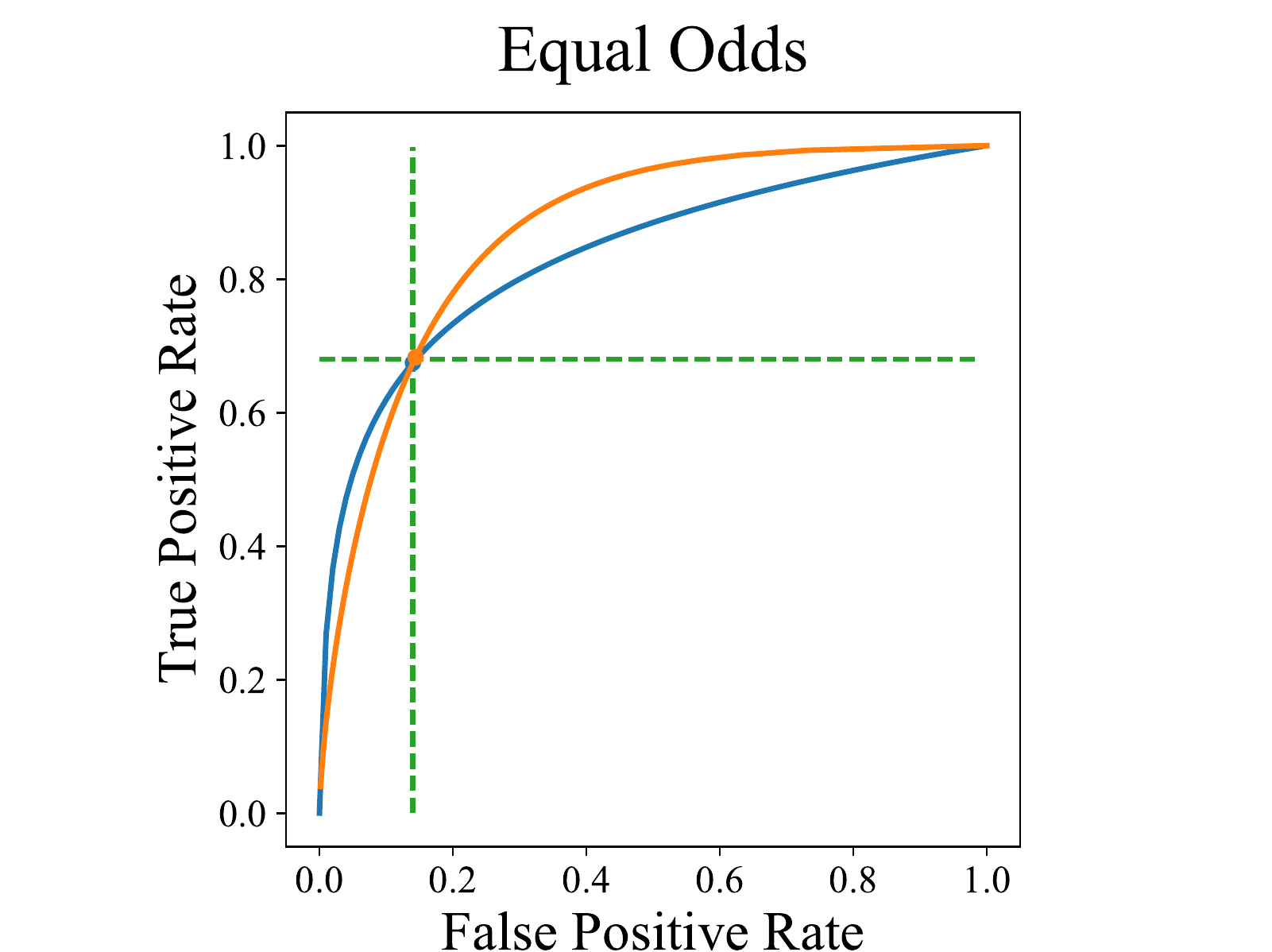}
\vspace{-0.5em}
\caption{Illustration of fairness goals forcing operating points on two groups' ROC curves.}
\label{fig:fairnessroc} 
\end{figure*}

\textbf{Accurate Coverage:} This fairness goal is often aligned with classification correctness. Consider for each group the true positive rate (TPR) and false positive rate (FPR), and the ratio of training data labelled positive (LPR) and negative (LNR). The goal of accurate coverage requires TPR x LPR + FPR x LNR = LPR for each group. As illustrated in Figure~\ref{fig:fairnessroc}-left, the resulting per-group constraint lines TPR + FPR x (LNR/LPR) = 1 cross the corresponding group's ROC curves to determine the operating point. Improvements in classifier accuracy can be captured while still satisfying this goal's constraints.

\textbf{Equal Opportunity:} This is a particularly challenging fairness goal since it requires the TPR to be the same for all groups, which forces each group's  operating point on its ROC curve to lie along some horizontal line (see Figure~\ref{fig:fairnessroc}-center).  Suppose one had achieved this goal (as pictured), but then we could improve the classifier for the orange group, resulting in its ROC curve moving towards the upper-left corner (TPR=1, FPR=0). The equal opportunity constraint might only be able to take advantage of the improved classifier's reduction in FPR, and not its improvement in TPR (though the best horizontal line could move up).

\textbf{Equal odds:} This goal is often hard to optimize for, and can only be achieved if all ROC curves pass through a single point (see Figure~\ref{fig:fairnessroc}-right), making it impractical for more than two protected groups. Forcing equal odds for more than two groups often requires intentionally degrading the classifier with stochasticity \citep{hardt:2016}, with the unappetizing cost of randomizing classification decisions.


\clearpage
\newpage


\begin{thebibliography}{20}
\providecommand{\natexlab}[1]{#1}
\providecommand{\url}[1]{\texttt{#1}}
\expandafter\ifx\csname urlstyle\endcsname\relax
  \providecommand{\doi}[1]{doi: #1}\else
  \providecommand{\doi}{doi: \begingroup \urlstyle{rm}\Url}\fi

\bibitem[Bauschke and Borwein(1996)]{bauschkeBorwein:96}
H.~Bauschke and J.~Borwein.
\newblock On projection algorithms for solving convex feasibility problems.
\newblock \emph{SIAM Review}, pages 367--426, 1996.

\bibitem[Bickel et~al.(1975)Bickel, Hammel, and {O'Connell}]{Bickel:75}
P.~J. Bickel, E.~A. Hammel, and J.~W. {O'Connell}.
\newblock Sex bias in graduate admissions: Data from {Berkeley}.
\newblock \emph{Science}, pages 398--404, 1975.

\bibitem[Calders and Verwer(2010)]{CaldersVerwer:2010}
T.~Calders and S.~Verwer.
\newblock Three naive {Bayes} approaches for discrimination-free
  classification.
\newblock \emph{Data Mining and Knowledge Discovery}, 2010.

\bibitem[Canini et~al.(2016)Canini, Cotter, Fard, Gupta, and
  Pfeifer]{canini:2016}
K.~Canini, A.~Cotter, M.~M. Fard, M.~R. Gupta, and J.~Pfeifer.
\newblock Fast and flexible monotonic functions with ensembles of lattices.
\newblock \emph{Advances in Neural Information Processing Systems {(NIPS)}},
  2016.

\bibitem[Dwork et~al.(2012)Dwork, Hardt, Pitassi, Reingold, and
  Zemel]{dwork:2012}
C.~Dwork, M.~Hardt, T.~Pitassi, O.~Reingold, and R.~Zemel.
\newblock Fairness through awareness.
\newblock In \emph{Proc. 3rd Innovations in Theoretical Computer Science},
  pages 214--226. ACM, 2012.

\bibitem[Fish et~al.(2016)Fish, Kun, and Lelkes]{fish:2016}
B.~Fish, J.~Kun, and A.~D. Lelkes.
\newblock A confidence-based approach for balancing fairness and accuracy.
\newblock \emph{Proc. {SIAM} Intl. Conf. Data Mining}, 2016.

\bibitem[Goh et~al.(2016)Goh, Cotter, Gupta, and Friedlander]{Goh:2016}
G.~Goh, A.~Cotter, M.~Gupta, and M.~P. Friedlander.
\newblock Satisfying real-world goals with dataset constraints.
\newblock In \emph{NIPS}, pages 2415--2423. 2016.

\bibitem[Gupta et~al.(2016)Gupta, Cotter, Pfeifer, Voevodski, Canini, Mangylov,
  Moczydlowski, and Esbroeck]{GuptaEtAl:2016}
M.~R. Gupta, A.~Cotter, J.~Pfeifer, K.~Voevodski, K.~Canini, A.~Mangylov,
  W.~Moczydlowski, and A.~V. Esbroeck.
\newblock Monotonic calibrated interpolated look-up tables.
\newblock \emph{Journal of Machine Learning Research}, 17\penalty0
  (109):\penalty0 1--47, 2016.
\newblock URL \url{http://jmlr.org/papers/v17/15-243.html}.

\bibitem[Haidt(2013)]{Haidt:2013}
J.~Haidt.
\newblock \emph{The Righteous Mind}.
\newblock Random House, New York, USA, 2013.

\bibitem[Hajian and {Domino-Ferrer}(2013)]{Hajian:2013}
S.~Hajian and J.~{Domino-Ferrer}.
\newblock A methodology for direct and indirect discrimination prevention in
  data mining.
\newblock \emph{{IEEE Trans. on Knowledge and Data Engineering}}, 25\penalty0
  (7):\penalty0 1445--1459, 2013.

\bibitem[Hardt et~al.(2016)Hardt, Price, and Srebro]{hardt:2016}
M.~Hardt, E.~Price, and N.~Srebro.
\newblock Equality of opportunity in supervised learning.
\newblock In \emph{NIPS}, 2016.

\bibitem[Kahneman(2011)]{kahneman:2011}
D.~Kahneman.
\newblock \emph{Thinking, Fast and Slow}.
\newblock Farrar Straus and Giroux, USA, 2011.

\bibitem[Kamiran et~al.(2010)Kamiran, Calders, and Pechenizkiy]{kamiran:2010}
F.~Kamiran, T.~Calders, and M.~Pechenizkiy.
\newblock Discrimination aware decision tree learning.
\newblock \emph{Discrimination aware decision tree learning}, pages 869--874,
  2010.

\bibitem[Kamishima et~al.(2012)Kamishima, Akaho, Asoh, and
  Sakuma]{kamishima:2012}
T.~Kamishima, S.~Akaho, H.~Asoh, and J.~Sakuma.
\newblock Fairness-aware classifier with prejudice remover regularizer.
\newblock \emph{Machine Learning and Knowledge Discovery in Databases}, pages
  35--50, 2012.

\bibitem[Kilbertus et~al.(2017)Kilbertus, Rojas~Carulla, Parascandolo, Hardt,
  Janzing, and Sch\"{o}lkopf]{Kilbertus:2017}
N.~Kilbertus, M.~Rojas~Carulla, G.~Parascandolo, M.~Hardt, D.~Janzing, and
  B.~Sch\"{o}lkopf.
\newblock Avoiding discrimination through causal reasoning.
\newblock In \emph{NIPS}, pages 656--666. 2017.

\bibitem[Lichman(2013)]{Lichman:2013}
M.~Lichman.
\newblock {UCI} machine learning repository, 2013.
\newblock URL \url{http://archive.ics.uci.edu/ml}.

\bibitem[Merton(2017)]{Merton:1968}
R.~K. Merton.
\newblock The {Matthew} effect in science.
\newblock \emph{{arxiv.org/pdf/1702.00502.pdf}}, 2017.

\bibitem[Russell et~al.(2017)Russell, Kusner, Loftus, and Silva]{Russell:2017}
C.~Russell, M.~J. Kusner, J.~R. Loftus, and R.~Silva.
\newblock When worlds collide: Integrating differetn counterfactual assumptions
  in fairness.
\newblock In \emph{NIPS}. 2017.

\bibitem[Thorndike(1920)]{thorndike:1920}
E.~L. Thorndike.
\newblock A constant error in psychological ratings.
\newblock pages 25--29, 1920.

\bibitem[Zafar et~al.(2017)Zafar, Valera, Rodriguez, and Gummadi]{Zafar:2017}
M.~B. Zafar, I.~Valera, M.~G. Rodriguez, and K.~P. Gummadi.
\newblock Fairness constraints: Mechanisms for fair classification.
\newblock In \emph{AIStats}, 2017.

\end{thebibliography}
\end{document}